\documentclass{article}
\usepackage{spconf,amsmath,graphicx}
\usepackage{graphicx}
\usepackage{comment}
\usepackage{amsmath,amssymb} % define this before the line numbering.
\usepackage{color}
\usepackage{amsmath}
\usepackage{amssymb}
\usepackage{subfigure}
\usepackage{booktabs} 
\usepackage{graphicx}
\usepackage[ruled]{algorithm2e}
\usepackage{multirow}
\newsavebox\CBox

\newtheorem{definition}{Definition} 

\newtheorem{proposition}{Proposition}
\newtheorem{proof}{Proof}

\title{InfoVAEGAN : learning joint interpretable representations by information maximization and maximum likelihood}
\name{Fei Ye and Adrian G. Bors}
\address{Department of Computer Science, University of York, York YO10 5GH, UK}
\begin{document}
%\ninept
%
\maketitle
\begin{abstract}
 Learning disentangled and interpretable representations is an important step towards accomplishing comprehensive data representations on the manifold. In this paper, we propose a novel representation learning algorithm which combines the inference abilities of Variational Autoencoders (VAE) with the generalization capability of Generative Adversarial Networks (GAN). The proposed model, called InfoVAEGAN, consists of three networks~: Encoder, Generator and Discriminator. InfoVAEGAN aims to jointly learn discrete and continuous interpretable representations in an unsupervised manner by using two different data-free log-likelihood functions onto the variables sampled from the generator's distribution. We propose a two-stage algorithm for optimizing the inference network separately from the generator training. Moreover, we enforce the learning of interpretable representations through the maximization of the mutual information between the existing latent variables and those created through generative and inference processes. 
\end{abstract}


\section{Introduction}

Unsupervised disentangled representation learning is a challenging task in any machine learning application. Most studies consider disentangled representation to be a data decomposition into sets of statistically and syntactically independent variables. Such data sets are assumed to be semantically distinct and to represent different categories of data characteristics. Learning disentangled representations that may capture semantic meaningful information can allow to explicitly edit images and is useful for a variety of tasks \cite{DeepVAEMixture,JointVAE_MI,LifelongVAEGAN}. Enabling disentangled representations can overcome overfitting during the training, leading to better generalization in models, \cite{InfoBottle}.

One of the most popular generative models is the Variational Autencoder \cite{VAE}, which implements a mapping between the data and an estimated latent space. The VAE's loss function maximizes the lower bound on the marginal log-likelihood of the data, while accurately reconstructing the data from the mapping of the latent space using the Kullback-Leibler (KL) divergence.
Learning interpretable and disentangled representations have been considered in $\beta-$VAE \cite{baeVAE} by setting a large penalty on the KL divergence term in order to encourage the independence between latent variables. On the other hand $\beta$-VAE sacrifices the quality of data reconstruction when inducing disentangled representations, \cite{DisentanglingByFactorising}. $\beta$-TCVAE model introduced the usage of the total correlation (TC) penalty,  which is a measure of multivariate mutual independence. The TC penalty was used in various VAE frameworks \cite{VAETCE} for inducing disentangled representations. However, TC is biased and is zero only if estimated on the whole dataset, \cite{Info_constraints_VAE}. Meanwhile, reducing the bias to zero is impossible for a large-scale dataset. The drawback of VAE based approaches is that they generally produce blurred and unclear images when compared to Generative Adversarial Networks (GANs) \cite{WGAN}. Few research efforts have been devoted to use GANs for disentangled representations \cite{Infogan}, and with mixed results. 

This research study has the following contributions~:
\vspace*{-0.1cm}
\begin{itemize}
	\item [1)] 
    A novel two-stage training algorithm where the inference model is estimated separately from the generator. 
    \vspace*{-0.6cm}
	\item [2)] 
	A data-free log-likelihood optimization approach able to learn an accurate inference model from a GAN.
\end{itemize}
 
\section{Background and related works}
\label{Back}

\noindent{\bfseries Variational autoencoder (VAE).} VAEs \cite{VAE} aim to maximize a lower bound to the marginal log-likelihood of the data~:
\begin{equation}
\begin{aligned}
\mathcal{L}(\phi,\theta)=& \mathbb{E}_{q_{\theta} ({\bf z}|{\bf x})}[\log p_{\phi} ({\bf x}|{\bf z})] -  D_{KL} (q_{\theta} ({\bf z}|{\bf x}) || p({\bf z}))  \\&\leq \log p({\bf x})
\end{aligned}
\label{eq1}
\end{equation}
where ${\bf x}$ and ${\bf z}$ are the input data and the corresponding latent variables, and the conditional distributions $q_{\theta}({\bf z}|{\bf x})$ and $p_{\phi}({\bf x}|{\bf z})$, are implemented by the Encoder and Decoder networks, of parameters $\theta$ and $\phi$, respectively. These networks are trained using the Stochastic Gradient Descent (SGD) algorithm. 

\noindent{\bfseries Generative adversarial networks (GAN).}
GANs also consist of two network components~: Generator and Discriminator which are trained for playing a Minimax game, defined by the following loss:
\begin{equation}
\begin{aligned}
\underset{G}{\mathop{\min }}\,\underset{D}{\mathop{\max }}\,V (D,G) &=
\mathbb{E}_{{\bf x}\sim{\ }pd ({\bf x})}[\log D({\bf x}) ]\\&+ \mathbb{E}_{{\bf z}\sim{\ }p ({\bf z})} [ \log [ 1-D ( G( {\bf z} )) ]].
\end{aligned}
\end{equation}
While the discriminator network is trained to distinguish between real and fake data, the generator aims to produce more realistic data that can fool the discriminator.
GANs are challenging to control and may generate unexpected results. 

\noindent{\bfseries Hybrid models. }
Hybrid models attempt to address the drawbacks of GANs and VAEs, by combining their architectures. These models usually have three components: an Encoder for mapping data into the latent space, a Generator to recover data from the latent space, and a Discriminator to distinguish real from fake data. Adversarial learning can be performed in the data space, latent space \cite{JointVAE_MI}, or on their joint spaces. 

Lately, the likelihood estimation as a regularization term was shown to stabilize adversarial distribution matching \cite{LifelongVAEGAN}. However, these methods only focus on improving the generation capability and do not design suitable objective functions for inducing disentangled representations. Our paper is the first to propose an appropriate objective function for training a hybrid VAE-GAN method for learning both continuous and discrete disentangled representations. 

\vspace*{-0.2cm}
\section{The InfoVAEGAN model}
\label{InfoVAEGAN}
\vspace*{-0.1cm}

The proposed InfoVAEGAN model is made up of three networks: Encoder, Generator and Discriminator.

\vspace*{-0.2cm}
\subsection{Generation from prior distributions}
\vspace*{-0.1cm}

Let ${\bf x} \in {\mathbb{R}^d}$ represent the observed random variables sampled from the empirical data distribution $\mathbb{P}_{\bf x}$. One of the goals of our model is to train the Generator to approximate the true data distribution $\mathbb{P}_{\bf x}$. Let us assume three underlying generative factors ${\bf z}, {\bf c}, {\bf d}$, corresponding to random, continuous and discrete variables, which are sampled from three independent prior distributions ${\bf z} \sim {\cal N} ({\bf I}^z, {\bf \Sigma}^z )$, ${\bf c} \sim {\cal N} ({\bf I}^c, {\bf \Sigma}^c )$, ${\bf d} \sim Cat (k=K ,p=1/K )$,  where $Cat$ denotes the Categorical distribution and ${\cal N}$ is he Gaussian distribution. Let us consider that the data generated ${\bf x}'$ is produced by a generator $G_\psi({\bf z},{\bf d},{\bf c})$, implemented by a neural network with trainable parameters $\psi$, and $\mathbb{P}_G$ to represent the distribution of data generated by $G$. The generation process is defined as:
${\bf d}\sim p({\bf d}),\;{\bf z}\sim p({\bf z}),c\sim p({\bf c}),{\bf x}\sim p_{\psi}({\bf x}|{\bf z},{\bf d},{\bf c})$.

 For the Discriminator network we use the Earth-mover distance, as in the Wasserstein GAN (WGAN) model \cite{WGAN}, which is defined as the optimal path of transporting information mass from the generator distribution $\mathbb{P}_G$ to the data distribution $\mathbb{P}_{\bf x}$. By considering the  Kantorovich-Rubinstein duality \cite{OptimalT}, the optimal transport adversarial learning is defined as:
\vspace{-3pt}
\begin{equation}
\begin{aligned}
\mathop {\min }\limits_G \mathop {\max }\limits_{D \in \Theta} \mathbb{E}_{{\bf x} \sim \mathbb{P}_{\bf x}} [ D ({\bf x}) ] - \mathbb{E}_{{\bf x}' \sim {\mathbb{P}_G}}  D ( {\bf x}') ]
\end{aligned}
\end{equation}
where $\Theta$ represents a set of 1-Lipschitz functions. We introduce a gradient penalty term \cite{ImprWGAN}, to enforce the Lipschitz constraint, resulting in:
\begin{equation}
\begin{aligned}
&\mathop {\min }\limits_G \mathop {\max } \limits_D 
 \mathbb{E}_{{\bf x} \sim {\mathbb{P}_{\bf x}}} [D ({\bf x}) ] - \mathbb{E}_{{\bf x}' \sim \mathbb{P}_G} [D({\bf x}')] \\&+ \lambda \mathbb{E}_{\tilde{\bf x} \sim \mathbb{P}_{\tilde{\bf x}}} [ ( \left\| {\nabla _{\tilde x}}
D ( \tilde{\bf x}) \right\|_2 - 1 )^2 ],
\end{aligned}
\end{equation}
where $\mathbb{P}_{\tilde{\bf x}}$ is defined as sampling uniformly along straight lines between pairs of data sampled from $\mathbb{P}_{\bf x}$ and $\mathbb{P}_G$.

\vspace*{-0.15cm}
\subsection{Data-free log-likelihood optimization}
\vspace*{-0.05cm}

In this section, we introduce two data-free log-likelihood optimization functions, which are used to learn the disentangled latent representations ${\bf z}$ and ${\bf u}=({\bf d},{\bf c})$, respectively. Instead of maximizing the sample log-likelihood, as commonly used in the VAE framework \cite{VAE}, we optimize the log-likelihood function by deriving a lower bound over the data samples drawn from the generator distribution. 

\begin{definition}
Let ${\bf x}' \sim G(\tilde{\bf z},\tilde{\bf d},\tilde{\bf c})$ be the generated data where $\tilde{\bf z},\tilde{\bf d},\tilde{\bf c}$ are latent variables sampled from the prior distributions $p(\tilde{\bf z})$, $p(\tilde{\bf d})$, $p(\tilde{\bf c})$.
\end{definition}

\begin{definition}
 Let $q_\omega({\bf d},{\bf c}|{\bf x}), q_\xi ({\bf z}|{\bf x})$ represent two independent conditional distributions implemented by two inference models. Let us define $({\bf d}, {\bf c})$ as interpretable representations which model discrete and continuous meaningful variations of the data and 
$\tilde{\bf z}$, ${\bf x}'$ as the observed variables. Let us define a latent variable model $p_\psi ( {\bf x}',\tilde{\bf z},{\bf d},{\bf c}) = p_\psi ( {\bf x}'|\tilde{\bf z},{\bf d},{\bf c}) p({\bf d},{\bf c}) p(\tilde{\bf z})$. Then, the log-likelihood of $p_\psi ( {\bf x}')$ is defined as:
\begin{equation}
\vspace*{-0.2cm}
\begin{aligned}
\log p_\psi ( {\bf x}') & =  \iiint \log p_\psi ( {\bf x}'|{\bf d},{\bf c},\tilde{\bf z} ) \: p ( {\bf d},{\bf c})\: p(\tilde{\bf z}) \: d {\bf d} \: d{\bf c} \: d \tilde{\bf z}.
 \end{aligned}
 \vspace*{-0.1cm}
\end{equation}
\end{definition}

This expression is intractable and can be rewritten by considering its Evidence Lower Bound (ELBO), as~:
\begin{equation}
\vspace*{-0.1cm}
\begin{aligned}
 \log p_\psi ( {\bf x}')  \ge 
 \mathbb{E}_{q_{\omega,\xi} ({\bf d}, {\bf c}, {\bf z}|{\bf x}')} \left[ \log \frac{p_\psi ( {\bf x}',
 \tilde{\bf d},\tilde{\bf c},{\bf z})}{q_\xi ( {\bf z}|{\bf x}') q_\omega ({\bf d},{\bf c}|{\bf x}')} \right].
 \label{logppsi}
 \end{aligned}
\end{equation}
The scheme for optimizing both $q_\omega ({\bf d},{\bf c}|{\bf x}')$ and $q_\xi ({\bf z}|{\bf x}')$, without updating the Generator, is very efficient.

\vspace*{-0.1cm}
\section{The theoretical framework}
\label{TheoFrame}
\vspace*{-0.1cm}

In existing hybrid methods, the inference model and the generator network are trained jointly by using a single objective function. However, in the proposed InfoVAEGAN model we have independent optimization procedures for the inference and generation. This choice has many advantages. For instance, the training of the inference model implemented by the Encoder, does not interfere with the optimization of the Generator, which results in a stable training procedure. When the Generator approximates the true data distribution exactly, we can derive more accurate inference models.  Aligning two joint distributions by using adversarial learning would also be harder to achieve than matching two single distributions individually. Unlike in InfoGAN \cite{Infogan}, the proposed model has a full inference mechanism, which enables the inference of both meaningful and nuisance latent representations, benefiting many down-stream tasks such as data reconstructions and interpolations.

\begin{proposition}
For a given inference model, we can estimate the testing data log-likelihood~:
\begin{equation}
\vspace*{-0.2cm}
\begin{aligned}
&  \log p_\psi ({\bf x}_t) \ge \mathbb{E}_{q_{\omega ,\xi} ( {\bf d},{\bf c},{\bf z}|{\bf x}_t)} [\log p_\psi ( {\bf d},{\bf c},{\bf z}|{\bf x}_t )] \\&- D_{KL} ( q_\omega ( {\bf d}|{\bf x}_t)||p ({\bf d})) - D_{KL} ( q_\omega ( {\bf c}|{\bf x}_t)||p ({\bf c}))  \\&- D_{KL} ( q_\xi ( {\bf z}|{\bf x}_t)||p({\bf z}) )=\mathcal{L} ( \psi ,\xi ,\omega ;{\bf x}_t)
\label{ELBO3_equation}
\end{aligned}
\end{equation}
where ${\bf x}_t$ represent testing data. The model implementing $p_\psi ({\bf x}_t)$ combines the two inference models and a Generator.
\end{proposition}

\begin{proof}
\vspace*{-0.1cm}
We combine the two inference models for continuous and discrete variables, and a Generator into a single model:
\begin{equation}
\log p_\psi ({\bf x}_t) = p_\psi ( {\bf x}_t|{\bf d,c,z} ) q_\omega ( {\bf d,c}|{\bf x}_t) q_\xi ( {\bf z}|{\bf x}_t)
\vspace*{-0.1cm}
\end{equation}
Then we define the model log-likelihood as~:
\begin{equation}
\vspace*{-0.3cm}
\log p_\psi ({\bf x}_t) = \log \mathbb{E}_{q_{\omega ,\xi} ( {\bf d,c,z}|{\bf x}_t)} \left[ \frac{p_\psi 
({\bf x}_t,{\bf d,c,z})}{q_{\omega ,\xi }( {\bf d,c,z}|{\bf x}_t} \right]
\end{equation}

According to the Jensen inequality, we have~:
\begin{equation}
\vspace*{-0.5cm}
\begin{aligned}
&\log p_\psi ( {\bf x}_t)  \ge \mathbb{E}_{q_{\omega ,\xi } ( {\bf d,c,z}|{\bf x}_t)} \left[ \log \frac{p_\psi ({\bf x}_t,{\bf d,c,z})}{q_{\omega ,\xi } ( {\bf d,c,z}|{\bf x}_t )} \right]\\
& = \mathbb{E}_{q_{\omega ,\xi } ( {\bf d,c,z}|{\bf x}_t)} \left[ \log \frac{p_\psi ( {\bf d,c,z}|{\bf x}_t) p ({\bf d}) p ({\bf c}) p({\bf z})}{q_\omega
( {\bf d}|{\bf x}_t) q_\omega ( {\bf c}|{\bf x}_t) q_\xi ( {\bf z}|{\bf x}_t )} \right]\\
&  = \mathbb{E}_{q_{\omega ,\xi } ( {\bf d,c,z}|{\bf x}_t)} \left[ \log p_\psi ( {\bf d,c,z}|{\bf x}_t) \right] - D_{KL} ( q_\omega  ( {\bf d}|{\bf x}_t )||p( {\bf d} ) ) \\&- D_{KL} ( q_\omega ( {\bf c|x}_t)||p({\bf c}) ) - D_{KL} ( q_\xi ( {\bf z|x}_t)||p ({\bf z} )).
 \end{aligned}
\end{equation}
\end{proof}

\begin{figure}[htbp]
\vspace*{-0.25cm}
	\centering
	\subfigure[Generator.]{
		\includegraphics[scale=0.54]{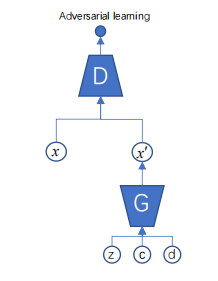}
	}
	\subfigure[Inference models.]{
		\includegraphics[scale=0.54]{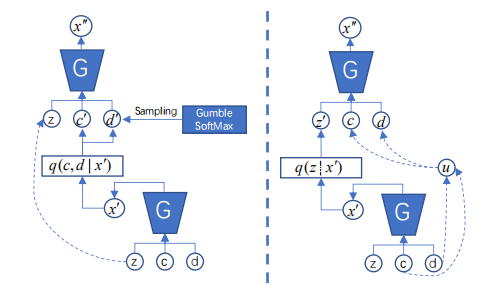}
	}
	\centering
	\vspace*{-0.3cm}
	\caption{Unsupervised learning structures in generative models, where ${\bf c}$ and ${\bf d}$ are continuous and discrete variables, while ${\bf z}$ represents Gaussian noise. }
	\label{fig1}
	\vspace{-5pt}
\end{figure}

\vspace*{-0.25cm}
\section{Mutual information maximization for interpretable representations}
\vspace*{-0.1cm}

 In the proposed InfoVAEGAN model, we transfer the underlying characteristic information of continuous and discrete latent variables during the decoder-generation process by using the Mutual Information (MI) maximization. Let us denote the joint latent variables by ${\bf u} = ({\bf d},{\bf c})$, while we want to maximize the MI between the joint latent variable ${\bf u}$ and the decoder output, $\rm{I}({\bf u},G({\bf z},{\bf u}))$. According to the research study from \cite{LifelongInterpretable} it is difficult to optimize the mutual information directly, given that it needs to access the information represented by the true posterior $p({\bf u}|{\bf x})$. In order to address this problem, we define an auxiliary distribution $W({\bf u}|{\bf x})$ to approximate the true posterior and then derive a lower bound on the mutual information, expressed by using the marginal entropy $H({\bf u})$, and the conditional entropy $H({\bf u}|G({\bf z},{\bf u}))$~:
\begin{equation}
\begin{aligned}
&{\rm{I}}({\bf u},  G({\bf z},{\bf u})) = H({\bf u}) - H({\bf u}|G({\bf z},{\bf u}))  \\
= & \iint G({\bf z},{\bf u}) p({\bf u}|{\bf x}) \log \frac{p({\bf u}|{\bf x})}{W({\bf u}|{\bf x})} \,d{\bf x}d{\bf u}  \\
&  + \iint G({\bf z},{\bf u})p({\bf u}|{\bf x}) 
\log W({\bf u}|{\bf x})d{\bf x}d{\bf u} + H({\bf u}) \vspace*{-4pt} \\
 =  & \;\; \mathbb{E}_{{\bf x} \sim G({\bf z},{\bf u})} 
D_{KL} [ p({\bf u}|{\bf x}) || W({\bf u}|{\bf x})] \\
&+ \mathbb{E}_{{\bf x} \sim G({\bf z},{\bf u})}
[ \mathbb{E}_{{\bf u} \sim p({\bf u},{\bf x})} 
[ \log W({\bf u}|{\bf x})] ] + H({\bf u})\\
 \geqslant & \mathbb{E}_{{\bf x} \sim G({\bf z},{\bf u})} 
[ \mathbb{E}_{{\bf u} \sim p({\bf u},{\bf x})} [ \log W({\bf u}|{\bf x})]] + H({\bf u}) = \mathcal{L}_{MI} 
\label{LMI}
\end{aligned}
\end{equation}
where the auxiliary distribution $W({\bf u}|{\bf x})$ is implemented by the Encoder. In practice, we sample a pair of latent variables ${\bf d}, {\bf c}$ from $q_\omega ( {\bf d},{\bf c}|{\bf x})$. We estimate the mutual information by means of the lower bound $\mathcal{L}_{MI}$, from (\ref{LMI}), while the last term $H({\bf u})$ represents the marginal entropy of the latent variables. 

The graph structure of the InfoVAEGAN is shown in Fig.~\ref{fig1}, where $q_\omega ({\bf d}|{\bf x})$ and $q_\omega ({\bf c}|{\bf x})$ are implemented by the same network except for the last layer which is different for the inference of each latent variable. The inference network, representing $q_\xi ({\bf z}|{\bf x})$, is implemented by a neural network with trainable parameters $\xi$, as it can be seen in the lower part of the left side of Fig.~\ref{fig1}b. The Generator is shown in Fig.~\ref{fig1}a.

\begin{figure}[htbp]
\vspace*{-0.15cm}
	\centering
    \includegraphics[scale=0.54]{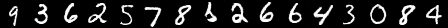}	
    \includegraphics[scale=0.54]{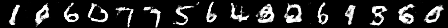}	
    \includegraphics[scale=0.54]{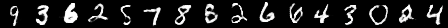}	
    \includegraphics[scale=0.54]{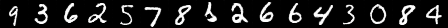}
    \vspace{-20pt}	
    \caption{Reconstruction results on each row: real images, reconstructions by ALI \cite{AdLearnInf}, InfoGAN \cite{Infogan} and InfoVAEGAN. }
    \vspace{-5pt}
    \label{Reconst}
\end{figure}

\hspace*{-3cm}
\begin{figure*}[htbp]
\vspace*{-0.2cm}
	\centering
	\begin{minipage}{0.31\linewidth}
	\includegraphics[scale=0.27]{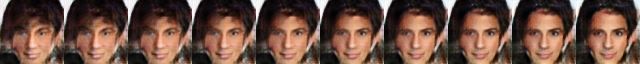}
	\includegraphics[scale=0.27]{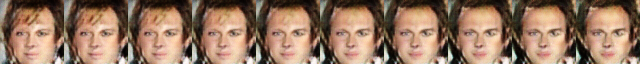}
	\centering
	\\
	\textbf{(a) Bangs} \\
	\includegraphics[scale=0.27]{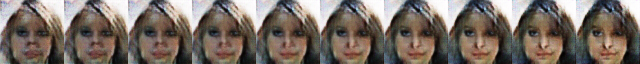}
	\includegraphics[scale=0.27]{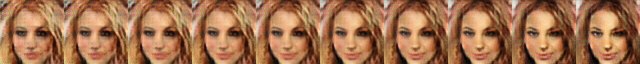}
	\centering
	\\
	\textbf{(c) Hair color} 
    \end{minipage} \hspace*{0.5cm}
\begin{minipage}{0.31\linewidth}
	\includegraphics[scale=0.27]{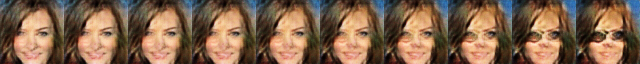}
	\includegraphics[scale=0.27]{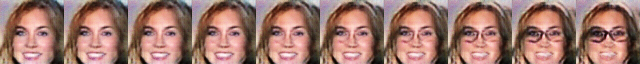}
	\centering
	\\
	\textbf{(b) Glasses} \\ 
	\includegraphics[scale=0.27]{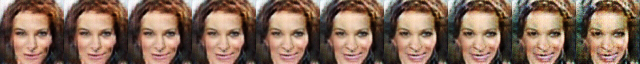}
	\includegraphics[scale=0.27]{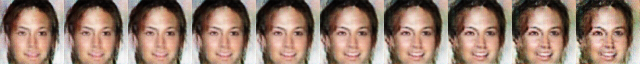}
	\centering
	\\
	\textbf{(d) Smile}
\end{minipage} \hspace*{0.9cm}
\begin{minipage}{0.27\linewidth}
      \includegraphics[scale=0.6]{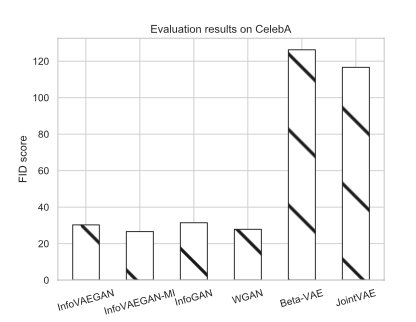} 
      \centering
	\\
	\textbf{(e) FID evaluation.}
\end{minipage}
\vspace*{-0.2cm}
	\caption{We change a single latent variable in the latent space from -1 to 1 while fixing all other latent variables for CelebA dataset in (a)-(d). FID evaluation when using CelebA database for training is provided in (e).}
	\label{Celeba_con}
	\vspace{-8pt}
\end{figure*}

\vspace*{-0.7cm}
\section{Experimental results}
\label{Exper}
\vspace*{-0.1cm}

In the following we evaluate the performance of InfoVAEGAN on the MNIST dataset \cite{lecun1998gradient}, representing images of handwritten digits. In order to learn the discrete latent variable which captures different styles of handwritten digits we use a categorical vector sampled from $Cat(K=10,p=0.1)$ and two continuous variables, sampled from the uniform distribution $U(-1,1)$, as latent variables.
The reconstruction results for the images from MNIST, shown in the first row from Fig.~\ref{Reconst}, by ALI  \cite{AdLearnInf}, InfoGAN \cite{Infogan}, and InfoVAEGAN, are provided in the following rows of images, respectively.
For the proposed InfoVAEGAN approach, the discrete latent variables are sampled from the Gumble-softmax distribution, while the continuous latent variables are sampled from the Gaussian distribution, whose mean and diagonal covariance are parameterized by the Encoder. From these results it can be observed that InfoVAEGAN provides better digit image reconstructions than InfoGAN or ALI.

\begin{figure}[htbp]
 \begin{center}
 \vspace*{-4pt}
 	\subfigure[InfoVAEGAN changing $c_1$.]{
		\includegraphics[scale=0.41]{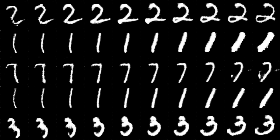}
	}
	\subfigure[InfoGAN changing $c_1$.]{
		\includegraphics[scale=0.41]{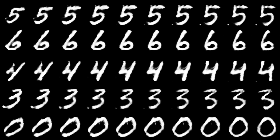}
	}
	\subfigure[InfoVAEGAN changing $c_2$.]{
		\includegraphics[scale=0.41]{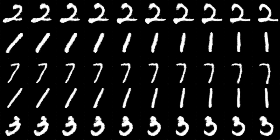}
	}
	\subfigure[InfoGAN changing $c_2$.]{
		\includegraphics[scale=0.41]{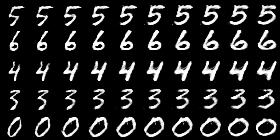}
	}
	\vspace{-20pt}
	\end{center}
	\caption{Generation results when changing the continuous variables $c_1$ and $c_2$ from -1 to 1. }
	\label{fig4}
	\vspace{-10pt}
\end{figure}

We modify the continuous codes $c_1$, $c_2$ within the range $[-1,1]$ and fix the other latent variables. The generative results for MNIST dataset are shown in Figures~\ref{fig4}a and \ref{fig4}c for InfoVAEGAN, while for InfoGAN are provided in Figures~\ref{fig4}b and \ref{fig4}d, when modifying $c_1$ and $c_2$. It can be observed that by varying the latent codes in InfoVAEGAN, we generate images showing meaningful characteristics such as rotations or a variety of handwriting styles. 
We also consider a 10-dimensional vector for the discrete and continuous latent variables in order to model underlying changing factors in the CelebA dataset \cite{Celeba}.  We change a single latent variable in the images generated by InfoVAEGAN while fixing the others. The results shown in Figures~\ref{Celeba_con}a-d indicate variations in face image representations such as bangs, glasses, hair colour and in smiling. 

The results when using InfoVAEGAN in unsupervised classification on the MNIST dataset, when compared with other methods, are provided in Table~\ref{Tab1}. Most unsupervised learning methods adopt mixture deep learning models ($K$ represents the number of components) requiring significantly more parameters. It observes that InfoVAEGAN achieves higher accuracy than InfoGAN \cite{Infogan}, and other models.

\begin{table}[]
\vspace*{-0.5cm}
    \centering
	\caption{Unsupervised classification results for $M$ runs.}
	\vspace*{0.2cm}
    \begin{tabular}{lcccc}
			\toprule
			\multicolumn{5}{c}{MNIST}                   \\
			\cmidrule(r){1-5}
			Method     & K & M    &  Mean    & Best   \\
			\midrule
			InfoVAEGAN     & 1 &4&95.42& 96.15\\
			JointVAE  \cite{JVAE}  & 1 &4&71.53&	87.32 \\
			SubGAN  \cite{subGAN}  & 20 &1&/&	90.81 \\
			InfoGAN  \cite{Infogan}  & 1 &1&/&	93.35 \\
			GMVAE \cite{Cluster_VAE}& 30 &1&/&	89.27 \\
		GMVAE \cite{Cluster_VAE}& 16 &1&/&	87.82 \\
			AAE \cite{AAE}& 16 &1&/&	90.45 \\
			CatGAN \cite{CatGAN}& 30 &1&/&	95.73 \\
			DEC \cite{DEC}& 10 &1&/&	84.30 \\
			PixelGAN \cite{pixelGAN_autoencoder}& 30 &1&/&	94.73 \\
		\end{tabular}
		\label{Tab1}
		\vspace*{-0.5cm}
\end{table}

We investigate the disentanglement ability of the proposed approach by using the metric from \cite{DisentanglingByFactorising} and the dataset dSprites \cite{dsprites}. The results are reported in Table~\ref{disentangled_tab}, where all other results are cited from \cite{JVAE}. The proposed approach achieves a competitive disentanglement score when compared with the current state of the art. We also use the Fréchet Inception Distance (FID) \cite{FID} to evaluate the quality of the generated images when considering the CelebA dataset in Fig.~\ref{Celeba_con}e, where InfoVAEGAN-MI denotes that the proposed approach does not use the mutual information (MI) loss. These results show that the proposed approach can balance well the disentanglement ability and image generation quality.

\begin{table}[]
    \centering
    \caption{Disentanglement evaluation on the dSprites.}
    \vspace*{0.2cm}
    \begin{tabular}{lll}
			\toprule
			\cmidrule(r){1-3}
			Methods   &M  &  Score    \\
			\midrule	InfoVAEGAN & 10 & 0.79 \\
			Beta-VAE \cite{baeVAE}  & 10 & 0.73 \\	FactorVAE \cite{DisentanglingByFactorising}&10 & 0.82 \\	JointVAE \cite{JVAE} & 10 & 0.69 \\
			\bottomrule
		\end{tabular}
		\label{disentangled_tab}
		\vspace{-15pt}
\end{table}

\vspace{-0.2cm}
\section{Conclusion}
\label{Con}
\vspace{-0.2cm}
	
In this paper, we introduce  InfoVAEGAN, a new deep learning approach for learning jointly discrete and continuous interpretable representations. InfoVAEGAN optimizes separately the inference model and the generator providing advantages over other hybrid methods. The proposed approach is a good tool to provide inference mechanisms when considering any generative GAN model without the need of any real data. In addition, InfoVAEGAN can generate high-quality interpretable data variations  which can successfully be used for disentangled and interpretable representation learning.

% References should be produced using the bibtex program from suitable
% BiBTeX files (here: strings, refs, manuals). The IEEEbib.bst bibliography
% style file from IEEE produces unsorted bibliography list.
% -------------------------------------------------------------------------
\bibliographystyle{IEEEbib}
\bibliography{VAEGAN.bib}

\end{document}